\documentclass[10pt,twocolumn,letterpaper]{article}

\usepackage{wacv}
\usepackage{times}
\usepackage{epsfig}
\usepackage{graphicx}
\usepackage{amsmath}
\usepackage{amssymb}
\usepackage{subcaption}

\def\wacvPaperID{6} 

\wacvfinalcopy 

\ifwacvfinal
\fi

\ifwacvfinal
\usepackage[breaklinks=true,bookmarks=false]{hyperref}
\else
\usepackage[pagebackref=true,breaklinks=true,colorlinks,bookmarks=false]{hyperref}
\fi

\ifwacvfinal
\else
\fi

\usepackage[absolute]{textpos}

\begin{document}

\begin{textblock*}{160mm}(1in,20pt)
\scriptsize{\noindent\copyright 2021 IEEE. Personal use of this material is permitted. Permission from IEEE must be obtained for all other uses, in any current or future media, including reprinting/republishing this material for advertising or promotional purposes, creating new collective works, for resale or redistribution to servers or lists, or reuse of any copyrighted component of this work in other works.}
\end{textblock*}
\begin{textblock*}{160mm}(1in,44pt)
\scriptsize{\noindent This is the author's version of a work that was accepted for publication in Proceedings of the 2021 IEEE Winter Applications of Computer Vision Workshops (WACVW). }
\end{textblock*}

\title{Focused LRP: Explainable AI for Face Morphing Attack Detection}

\author{Clemens Seibold\\
Fraunhofer HHI\\
{\tt\small clemens.seibold@hhi.fraunhofer.de}
\and
Anna Hilsmann\\
Fraunhofer HHI\\
{\tt\small anna.hilsmann@hhi.fraunhofer.de}
\and
Peter Eisert\\
Humboldt University Berlin \&\\ Fraunhofer HHI\\
{\tt\small peter.eisert@hu-berlin.de}
}

\maketitle
\thispagestyle{empty}

\begin{abstract}
The task of detecting morphed face images has become highly relevant in recent years to ensure the security of automatic verification systems based on facial images, e.g. automated border control gates. Detection methods based on Deep Neural Networks (DNN) have been shown to be very suitable to this end. However, they do not provide transparency in the decision making and it is not clear how they distinguish between genuine and morphed face images. This is particularly relevant for systems intended to assist a human operator, who should be able to understand the reasoning.
In this paper, we tackle this problem and present Focused Layer-wise Relevance Propagation (FLRP). This framework explains to a human inspector on a precise pixel level, which image regions are used by a Deep Neural Network to distinguish between a genuine and a morphed face image. Additionally, we propose another framework to objectively analyze the quality of our method and compare FLRP to other DNN interpretability methods. This evaluation framework is based on removing detected artifacts and analyzing the influence of these changes on the decision of the DNN. Especially, if the DNN is uncertain in its decision or even incorrect, FLRP performs much better in highlighting visible artifacts compared to other methods.
\end{abstract}

\section{Introduction}
A morphed face image is a synthetically generated face image that resembles two different subjects. The similarity is so strong that even biometric verification systems match the face of both subjects with this synthetic image. If such a picture was to be used in an identification document, two subjects could claim the ownership of this document and thus share one identity. Ferarra et al.~\cite{Ferrara14} raised awareness about this problem and its consequences for automatic face verification, especially for automatic border control systems.\\
The threat arising from morphed face images has prompted researchers to investigate this problem and to develop methods to detect them. A comprehensive overview on work on face morphing attacks and detectors can be found in \cite{Lit1, Lit2, Lit3}.\\ 
The existing detection methods can be divided into blind and non-blind face morphing attack detectors. The non-blind detectors make use of reference data, e.g.~a trusted image~\cite{Ferrara18, Scherhag20} or a 3D-model~\cite{Seibold18b} of the subject. In contrast to that, blind detectors use only the image that needs to be checked. Most of them are based on analysis of statistical characteristics such as image quality (e.g.~gradient distribution or spacial frequency distribution)~\cite{Neubert17,Neubert19}, Benford Features~\cite{MakrushinND17} or camera noise pattern~\cite{PRNUVA2}, using handcrafted features, or on statistical~\cite{Raghavendra16} or learned features~\cite{Raghavendra17}, e.g.~Deep Neural Networks (DNNs). While it is clear what information the methods based on handcrafted features use for their decision, this is not the case for methods based on learned features.
The later are, without further investigations, black-box detection systems: i.e. they are not transparent in their decision-making. Even though interpretability of DNNs is a prominent research topic, the studies about its applicability to DNN-based face morphing attack detectors are scarce. A recent research regarding this topic has been done by Seibold et al.~\cite{SeiboldSHE20}. They use Layer-wise Relevance Propagation (LRP)~\cite{BachPLOS15} to analyze on which coarse regions their DNN-based detectors focus and propose a training method that forces their DNNs to include information from all of these regions in the decision-making process. They also showed that in this case LRP cannot be used directly to understand this decision-making process without additional investigations.

\begin{figure*}
\begin{center}
\includegraphics[width=0.245\textwidth]{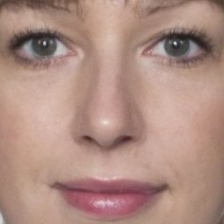}
\includegraphics[width=0.245\textwidth]{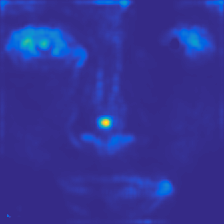}
\includegraphics[width=0.245\textwidth]{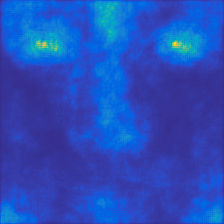}
\includegraphics[width=0.245\textwidth]{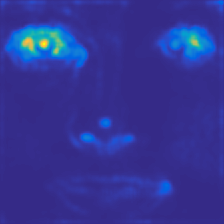}
\end{center}
\caption{Morphed face image (left) and visualization of different interpretability methods of a DNN-based face morphing detector. The interpretability methods LRP, Sensitivity Maps and FLRP (left to right) assign a relevance score to each pixel. A low score is visualized in blue and a large score in yellow. The input images for the morphed face image are from the Face Research Lab London Set~\cite{London}}
\label{fig:teaser}
\end{figure*}

In this paper, we present FLRP, which is an extension to LRP that determines which regions in an image can be used by a Deep Neural Network to distinguish between a genuine and a morphed face image. It focuses on large activations of neurons in the last layer of the feature extractor that are caused exclusively by morphed face images. It is intended to support a human in explaining why an image is a morphed face image by adding transparency to the decision-making process of DNN-based detectors. In contrast to LRP, it does not directly show which regions lead to a strong activation of a neuron that represents a class. Rather, it focuses on  neurons in an intermediate layer for which activation values are large, if the input image is a morphed face image, and the sources of these activations. In order to access whether our approach succeeds in detecting the relevant pixels, we propose an additional framework for evaluation of interpretability methods for DNN-based face morphing attack detectors. Finally, we compare FLRP with other interpretability methods for DNNs (Sensitivity Maps~\cite{SensitivityMaps} and LRP~\cite{BachPLOS15}). Figure \ref{fig:teaser} shows an example for the interpretability methods that are studied within this paper.\\

The key contributions of our paper are:
\begin{itemize}
\setlength\itemsep{0em}
\item We propose a new interpretability method for DNN-based blind Morphing Attack Detectors that precisely determines which regions of an image contain artifacts caused by face morphing.
\item We propose a new framework for evaluation of interpretability methods for Morphing Attack Detectors.
\item We evaluate our proposed interpretability method and show its advantages compared to other approaches when applied to Face Morphing Attack Detectors.
\end{itemize}

The structure of our paper is as follows. In the next section, we present the popular interpretability methods LRP and Sensitivity Maps and introduce FLRP.
Subsequently, we describe our framework for evaluation of interpretability methods for face morphing attack detectors based on DNNs in Section \ref{sec:framework}. 
The experimental setup for our evaluation of FLRP is described in Section \ref{sec:expSetup} and the results and a comparison of FLRP with LRP and Sensitivity Maps in Section \ref{sec:results}. We finish our paper with a summary and a discussion on further planed experiments and extensions to ensure the transparency and reliability of face morphing attack detectors based on DNNs.

\section{Interpretability of DNNs using Backward Propagation Techniques}
\label{sec:tech}
Most interpretability methods based on backward propagation assign a start value (relevance) to one neuron in the last layer, in which each neuron represents exactly one class. This relevance is then backpropagated into the input image based on the activations in the intermediate layers of the DNN and a method-dependent set of rules. Two very common methods that define how the relevance can be backpropagated into the input image are Sensitivity maps and LRP. In the following, we present the DNN used in our experiments, we briefly explain the concepts of Sensitivity Maps and LRP, and introduce FLRP.

\subsection{VGG-A}
\begin{figure}
\includegraphics[width=0.5\textwidth]{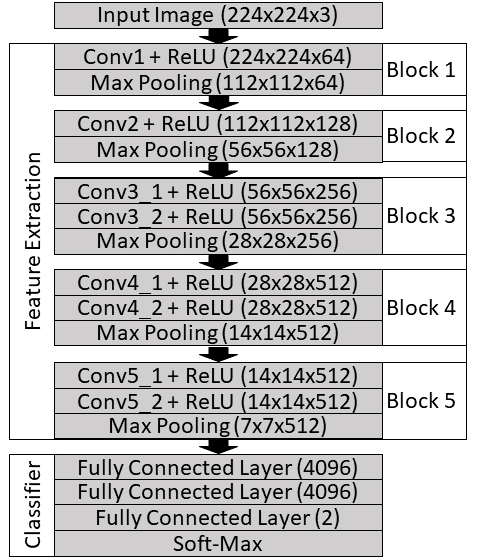}
\caption{VGG-A Architecture. The figures in brackets show the output size of the layer.}
\label{fig:VGGA}
\end{figure}

In our experiments, we use the DNN  architecture VGG-A~\cite{VGG} with an input size of $224\times224$ pixels. The VGG-A architecture follows the classical scheme for DNNs for image classification. It starts with blocks consisting of convolutional layers followed by a Rectified Linear Unit (ReLU), each of them ending with a max-pooling layer that reduces the spatial dimension. This part of the neural networks is also referred to as feature extractor. The feature extractor is followed by two fully connected layers, which are referred to as classification part of the neural network. Figure~\ref{fig:VGGA} shows the VGG-A architecture in more detail.

\subsection{Sensitivity Maps}
Sensitivity Maps~\cite{SensitivityMaps} are based on the partial derivatives of the activation function of a selected neuron with respect to the pixels in the image. They describe the influence of a change of a pixel on the change of the activation of the neuron that represents the selected class. This makes them a suitable means for analyzing the decision-making process of DNNs. The calculation of the Sensitivity Maps is done by using the concept of error backpropagation in the same way it is used for training of DNNs. 
A fictional error is defined for the class of interest and this error is backpropagated into the input image. The final relevance score for a pixel can now be calculated by applying a vector norm on the backpropagated errors of the color channels.\\
In this paper, we use the $L_1$ norm and apply it already after the second convolutional layer. In oder to propagate the relevance from this layer into the input image, it is uniformly distributed over all pixels that are reached  by the size of the convolutional filter. This approach is similar to our setting for LRP, prevents noise maps and leads to a more smoothed relevance distributions.

\subsection{Layer-wise Relevance Propagation}
While Sensitivity Maps answer the question of which region of the input images should be modified to maximally change the activation of a neuron that represents a class, LRP indicates what leads to and what inhibits the activation of this neuron. The activating relevance is also referred to as positive relevance and the inhibiting one as negative relevance. The theory behind LRP is based on a "deep Taylor decomposition" of the neural network function~\cite{BachPLOS15}. Similarly to the case of Sensitivity Maps, LRP assigns step-by-step (layer-by-layer) relevance from one selected neuron, which represents one class, through the DNN back to the image. For each layer, the relevance is backpropagated into the previous one using a set of rules. These rules are intended to direct the relevance towards the neurons in the previous layer that play an important role in the activation of each of the neurons in the current layer. It is possible to use LRP with different sets of rules for the relevance backpropagation. In this paper, we use the rules that are current best practice for LRP for similar structured DNNs~\cite{LRPRules}. These are: $\epsilon$-decomposition for the fully connected layers, $\alpha\-\beta$-decomposition (with $\alpha=2, \beta=-1$) for the convolutional layers in the blocks 3-5 (see Figure \ref{fig:VGGA}) and flat-decomposition for the first two convolutional layers. While the $\epsilon$-decomposition treats activating and inhibiting relevance similarly, the $\alpha\-\beta$-decomposition considers them separately and, with these particular values for $\alpha$ and $\beta$, focuses more on activating relevance, leading to more balanced results. The flat-decomposition propagates the relevance of a neuron equally distributed to all neurons in the previous layer that have an influence on this neuron. For a more detailed explanation of these methods, we refer to~\cite{LRPRules}.\\
\subsection{Focused LRP}
Our method, FLRP, is inspired by the results shown in \cite{SeiboldSHE20}. The authors used LRP to analyze on which coarse region a DNN for face morphing attack detection focuses for its decision-making. They showed that, due to the complexity of the behavior of the fully connected layers of a DNN, the relevance scores provided by LRP are not directly interpretable, requiring further investigations to understand the overall behavior of the network. They also showed that LRP often assigns large relevance scores to artifact-free regions in morphed face images, marking them as relevant for the decision of labeling the images as a morph.
In contrast to the studies in \cite{SeiboldSHE20}, which focused on average relevance distributions for a set of images, our study focuses on the independent processing of single images and is intended to provide more transparency for individual decisions of DNNs.\\

The main idea behind FLRP is to start the relevance propagation from neurons in an intermediate layer instead of those in the final one as it is the case in regular LRP, and thus to focus on the learned features that characterize a morphed face image. 
FLRP starts in the layer right before the classifier (i.e.~in the last max pooling layer) and assigns initial relevance scores to pre-selected neurons in this layer. In the following, we describe how these neurons are selected.
In our experiments, we use a VGG-A architecture with an input size of $224\times224$ pixels (see Figure \ref{fig:VGGA}). Thus, the last layer of the feature extractor has an output shape of $7\times7\times512$. Instead of a single neuron that represents a class, we have to assign an initial relevance score to these neurons of the feature extractor.
Since we are not interested in what leads to the activation of all of these neurons, but in what is typical for a morphed face image, we select a set of neurons that have strong activation values for morphed face images. It is expected that these will allow distinguishing between genuine and morphed face images. These neurons are selected based on the training data as described in the following.\\
In a first step, we calculate the output of the feature extraction part of the DNN for each image in the training data. This output consists of a $7\times7\times512$ tensor for each image. This tensor can be interpreted as an image with 512 channels and a size of $7\times7$ pixels. For each pixel, we select the channel that has a larger value when the input is a morphed face image and is best suited to distinguish between genuine and morphed face images. To this end, we calculate for each neuron a threshold such that the number of morphed face images that lead to activation values above this threshold is equal to the number of genuine face images that lead to activation values below that threshold. Based on these thresholds, we select for each pixel in the 512-channel "image" the channel that is most suitable to separate between genuine and morphed face images. This yields 49 $(7\times7)$ neurons, which we will use to initialize our relevance propagation. In contrast to common LRP or sensitivity maps, which we start from a single neuron and changing the starting value only scales the result, in FLRP it is necessary to assign suitable initial values for these neurons. To do so, we pass the image that should be inspected through the Neural Network and use the resulting activation values scaled based on the previously calculated equal error rates of the selected neurons as start relevance. The idea behind this initialization method is to assign start relevance mainly to neurons that did detect face morphing related artifacts and thus have large activation values. Starting with this assignment of relevance in the last layer of the feature extractor, we use the $\alpha\-\beta$ rule from LRP with $\alpha=2$  and $\beta=-1$  for all but the first two convolutional layers to propagate the relevance into the input image. For the first two convolutional layers, we use flat decomposition.

\section{Evaluation Framework for Interpretability Methods for DNN-based Face Morphing Attack Detectors}
\label{sec:framework}
In this section, we introduce our framework for evaluation of interpretability methods for DNNs-based morphing attack detectors. Our framework is designed for interpretability methods that assign a score to every pixel in the input image representing the estimated relevance for the decision of the DNN. This is a common approach to explain the decision-making of DNNs~\cite{BachPLOS15}, used by the majority of  interpretability methods that are based on backpropagation such as Sensitivity Maps or LRP.

\begin{figure}
\includegraphics[width=0.5\textwidth]{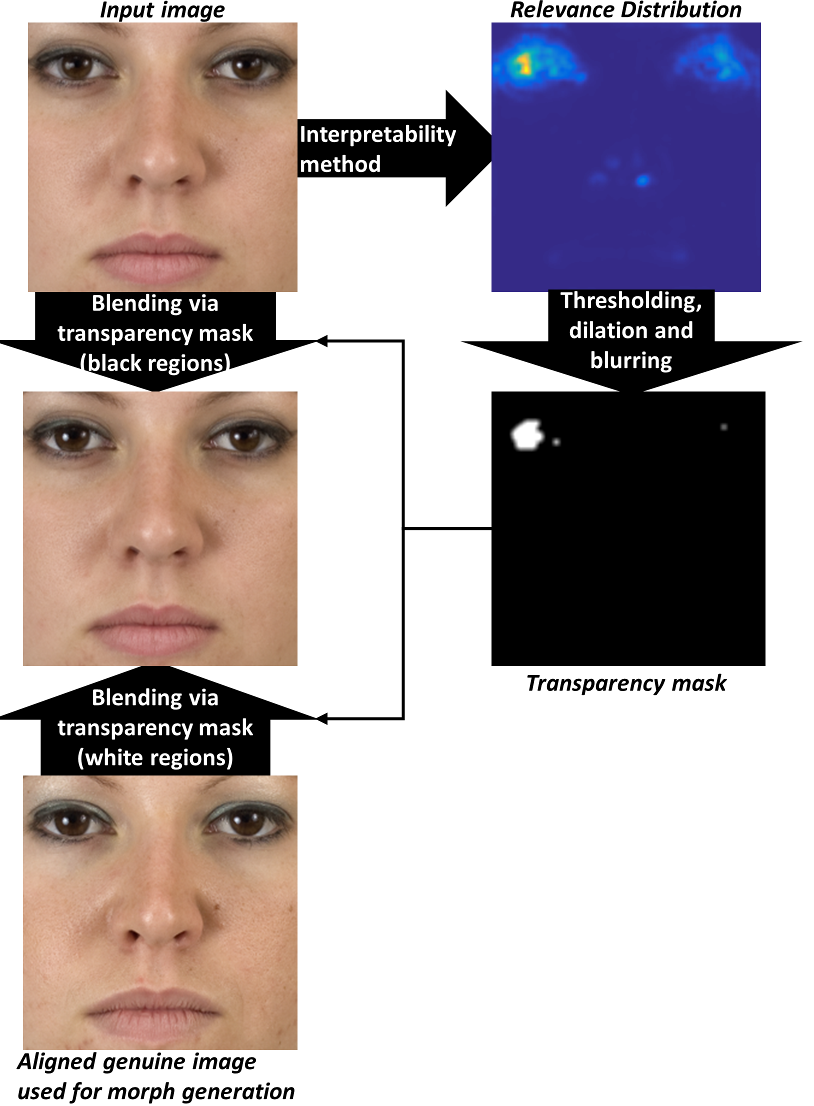}
\caption{Visualization of single steps of our interpretability evaluation method. First, the interpretability method assigns a relevance score to every pixel of the morphed face image. Based on these relevance scores, a transparency mask is generated such that the most relevant pixels have a transparency value of one (here shown in white). This mask is then blurred to ensure a smooth transition between the blended images. Finally, the morphed face image and an aligned genuine image are blended according to the transparency mask. If a pixel in the transparency mask has a value of one, the color value from the aligned genuine image is used.}
\label{fig:EvalMethod}
\end{figure}

The goal of our framework is to evaluate whether the pixels that have a high relevance score are actually relevant for the decision-making of the network and whether substituting these pixels changes the classification score. In contrast to \cite{Montavon2018} who also selected regions to be substituted by relevance and set these regions to a default color or random noise, we apply a more sophisticated substitution method. Setting regions to a default color or random noise would shift the image far off the distribution of aligned face images. Due to the binary nature of the task of face morphing detection, differences in the DNN predictions caused by such changes are not expected to convey meaningful information about the quality of the analyzed explainability methods. We assume that the relevant regions/pixels contain artifacts that are caused by the generation process of the morphed face images. Thus, removing the artifacts should change the decision of the network and be, therefore, an appropriate instrument for the evaluation of the relevance scores. In order to remove the possible artifacts, we substitute the regions that are marked as relevant in the morphed face image with the corresponding ones from the original image. Additionally, we smoothen the transition between the substituted part and the rest of the image. To this end, we use the following procedure:
\begin{enumerate}
\item For each generated morphed face image, we additionally store aligned input images. These aligned input images are usually generated as part of a face morphing pipeline, as it is the case in this study (see \cite{Seibold17} for alignment in face morphing pipelines).
\item We create a binary mask that describes which regions should be substituted. This mask contains $\alpha\%$ of the pixels with the largest relevance score according to the analyzed interpretability method.
\item We dilate the mask using a filter with a size of $3\times3$.
\item The binary mask is converted to a transparency-mask and blurred using a $5 \times 5$ blur kernel.
\item Based on the transparency-mask, we blend in corresponding regions from the aligned input images.
\end{enumerate}
The blurring of the mask ensures a smooth transition between the inserted part and the rest of the morphed face image. Moreover, due to the dilation the relevant pixels are completely substituted despite of the blurring. For the evaluation of the interpretability methods, we do not use a fixed $\alpha$, but analyze the change of the DNN's decision and loss with respect to $\alpha$.
Figure \ref{fig:EvalMethod} visualized this proposed algorithm.

\begin{figure*}[htb]
\includegraphics[width=\textwidth]{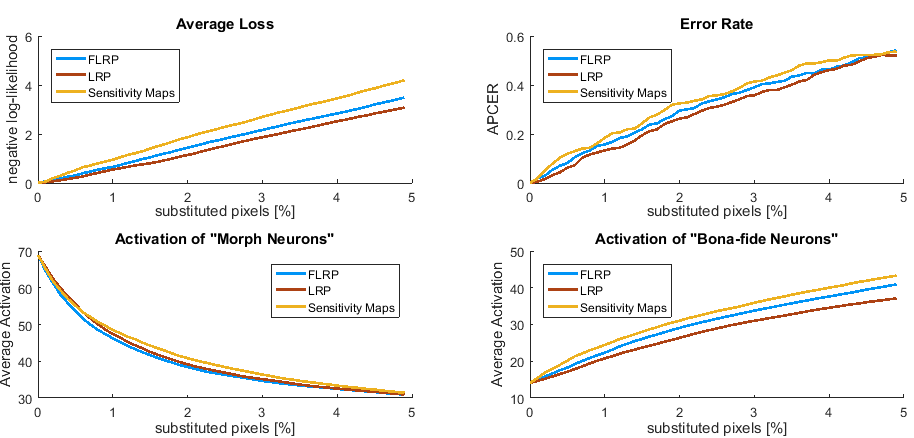}
\caption{Change of average loss, error rates, activation of selected neurons that have a large activation if the image is a bona-fide or morphed with respect to the amount of substituted pixels. The pixels are substituted according to our framework described in Section \ref{sec:framework} and the relevance distribution based on LRP, Sensitivity Maps and FLRP.}
\label{fig:stat}
\end{figure*}

\section{Experimental Setup}
\label{sec:expSetup}
\subsection{Data}
For our experiments, we have collected genuine face images from different public available datasets~\cite{CFD,London,BU4DFE,PUT,utrecht,scFace, FERET} and from our internal datasets, yielding about 2,000 genuine face images. We have  generated the same amount of morphed face images using an automatic pipeline described in \cite{SeiboldSHE20}. We split the images into a training set (70\%), a testing set (20\%), and a validation set (10\%). When splitting the genuine and morphed face images into these sets, we ensure that a subject and all morphed face images based on this subject are always in the same set and only in that one. 
\subsection{Detector and Performance}
For our experiments, we use a DNN-based detector similar to the one described in \cite{Seibold17}. It consists of a pre-processing step to normalize the data and a DNN that classifies the normalized input image. In order to normalize the image, we estimate facial landmarks, crop the inner part of the face and resize it to $224\times224$ pixels. As previously mentioned, we use the VGG-A architecture as our DNN and start our training with a model pre-trained for object classification on the ImageNet dataset. During the training, we used different image augmentation techniques, similarly to \cite{SeiboldSHE20}. We applied random jittering in the range of [-2, +2] pixels, random flipping, Gaussian and Salt-and-Pepper Noise and Gaussian and motion blur.\\
The performance of our detector is reported in table \ref{tab:performance} using the metrics Attack Presentation Classification Error Rate (APCER),  Bona-fide Presentation Classification Error Rate (BPCER) and Equal-Error-Rate (EER). APCER and BPCER are designed for the evaluation of detection systems for presentation attacks~\cite{PAISO}, but can directly be adapted for evaluation of the face morphing attack detectors. APCER is defined as the relative amount of attacks that are not detected and BPCER as the relative amount of genuine face images that are falsely classified as morphed face images.
\begin{table}[!h]
\begin{center}
\begin{tabular}{ c | c | c}
APCER & BPCER & EER\\
\hline
4.9\% & 2.6\% & 3.3\%
\end{tabular}
\end{center}
\caption{Performance of our DNN-based face morphing attack detector}
\label{tab:performance}
\end{table}

\section{Results}
\label{sec:results}
\begin{figure*}[htb]
\includegraphics[width=\textwidth]{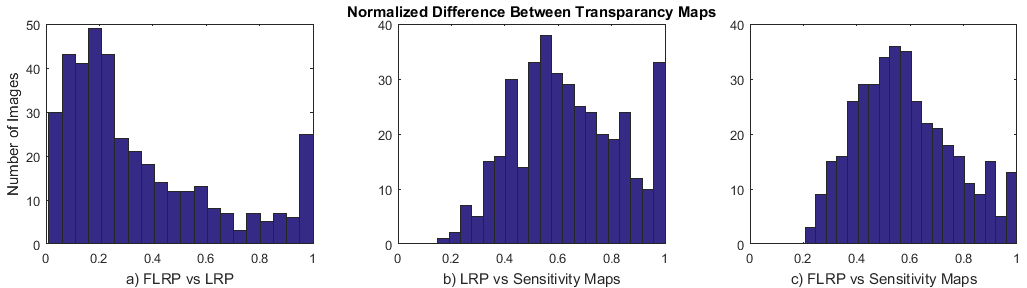}
\caption{Histograms of cumulative differences between the transparency maps derived from different interpretability methods.}
\label{fig:hmComparison}
\end{figure*}

In this section, we analyze the performance of FLRP using our proposed evaluation framework on the test data. Additionally, we compare FLRP to LRP and Sensitivity Maps. We use the change of the DNN's decision and loss (negative log-likelihood) with respect to the amount of substituted pixels in a morphed face image as performance metrics. It should be noticed that it is not possible to have ground truth data to compare our results with, since we cannot know how many and which regions need to be exchanged in the best case. However, we can compare the methods to each other. It is expected that the change of loss and APCER should be proportional to the quality and accuracy of the relevance predictions of an interpretability method. In addition, we report the average activation of the neurons that were chosen for the initialization of FLRP (in the following referred to as "Morph Neurons"). The activations of the similarly chosen set of neurons that are activated by genuine face images (later referred to as "Bona-fide Neurons") are also reported.\\

\subsection{General Performance}
The evaluation is performed on the morphed face images from the test data. Morphs that are not detected have been removed, since our general performance analysis focuses on the explainability of the causes that lead to a detection. However, they are included in the further analysis in the next subsection, which studies the cases in which the relevance distribution between FLRP and LRP strongly differs. Figure \ref{fig:stat} shows the average loss and APCER with respect to the amount of substituted pixels ($\alpha$) in the morphed face images for the different methods. The Sensitivity Maps method outperforms LRP and FLRP in terms of Average Loss and Error Rate changes. The performance of FLRP is between that of LRP and Sensitivity Maps.\\
In order to gain a better understanding about what differs between the regions selected by Sensitivity Maps and the regions selected by FLRP/LRP, we need to analyze the activation of Morph Neurons and Bona-fide Neurons. The average activation of the Morph Neurons decreases most for FLRP and least for Sensitivity Maps and the average activation of the Bona-fide Neurons increases most for Sensitivity Maps and least for LRP. Even though the selection based on Sensitivity Maps is most suitable to change the decision and loss of the DNN, it does not seem to lead to the strongest deactivation of neurons that are indicating artifacts. This agrees with the previously mentioned hypothesis that Sensitivity Maps focus on what leads to a change of the decision, not necessarily pointing to the artifacts typical for morphed face images, whereas LRP shows which regions are relevant for the activation of neurons. We believe that focusing on the morphing artifacts could be more useful when supporting the decisions of a human operator.\\
As a summary, LRP, Sensitivity Maps and FLRP seem to be quite similar when it comes to determining regions that are relevant for the detection of morphed face images, but FLRP focuses most on artifacts that lead to an activation of the Morph Neurons. It should also be mentioned that FLRP is more suitable compared to LRP to determine the regions that change the output of the network, even though it does not consider the network's classifier.

\subsection{FLRP for Uncertain Decisions}
\begin{figure}
\begin{center}
\begin{subfigure}[b]{0.23\textwidth}
\includegraphics[width=\textwidth]{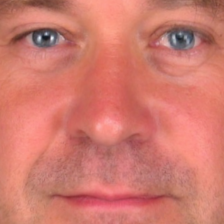}
\centering
{\small (a) input image}
\end{subfigure}
\begin{subfigure}[b]{0.23\textwidth}
\includegraphics[width=1\textwidth]{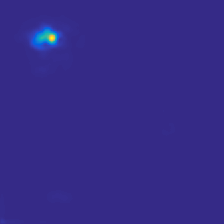}
\centering
{\small (b) LRP}
\end{subfigure}
\begin{subfigure}[b]{0.23\textwidth}
\includegraphics[width=\textwidth]{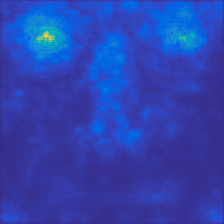}
\centering
{\small (c) Sensitivity Map}
\end{subfigure}
\begin{subfigure}[b]{0.23\textwidth}
\includegraphics[width=1\textwidth]{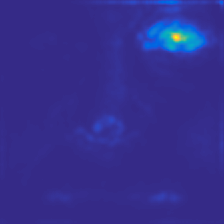}
\centering
{\small (d) FLRP}
\end{subfigure}
\begin{subfigure}[b]{0.23\textwidth}
\includegraphics[width=\textwidth]{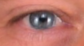}
\centering
{\small (e) magnified right eye}
\end{subfigure}
\begin{subfigure}[b]{0.23\textwidth}
\includegraphics[width=1\textwidth]{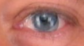}
\centering
{\small (f) magnified left eye}
\end{subfigure}
\end{center}
\caption{Example of relevance distributions (b)-(d) for a morphed face image (a) that was not detected as morph by the DNN.}
\label{fig:UseCaseExample1}
\end{figure}

\begin{figure}
\begin{center}
\begin{subfigure}[b]{0.23\textwidth}
\includegraphics[width=\textwidth]{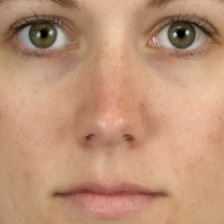}
\centering
{\small (a) input image}
\end{subfigure}
\begin{subfigure}[b]{0.23\textwidth}
\includegraphics[width=1\textwidth]{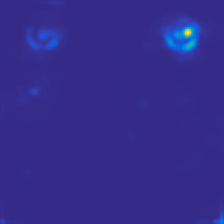}
\centering
{\small (b) LRP}
\end{subfigure}

\begin{subfigure}[b]{0.23\textwidth}
\includegraphics[width=\textwidth]{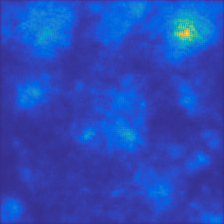}
\centering
{\small (c) Sensitivity Map}
\end{subfigure}
\begin{subfigure}[b]{0.23\textwidth}
\includegraphics[width=1\textwidth]{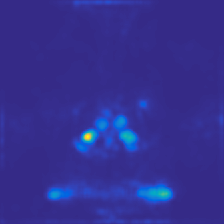}
\centering
{\small (d) FLRP}
\end{subfigure}
\begin{subfigure}[b]{0.23\textwidth}
\includegraphics[width=\textwidth]{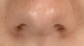}
\centering
{\small (e) magnified nose}
\end{subfigure}
\begin{subfigure}[b]{0.23\textwidth}
\includegraphics[width=1\textwidth]{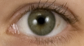}
\centering
{\small (f) magnified left eye}
\end{subfigure}
\end{center}
\caption{Example of relevance distributions (b)-(d) for a morphed face image (a) that was not detected as morph by the DNN.}
\label{fig:UseCaseExample2}
\end{figure}

\begin{figure}
\begin{center}
\begin{subfigure}[b]{0.23\textwidth}
\includegraphics[width=\textwidth]{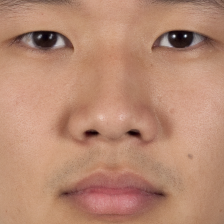}
\centering
{\small (a) input image}
\end{subfigure}
\begin{subfigure}[b]{0.23\textwidth}
\includegraphics[width=1\textwidth]{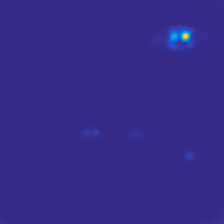}
\centering
{\small (b) LRP}
\end{subfigure}

\begin{subfigure}[b]{0.23\textwidth}
\includegraphics[width=\textwidth]{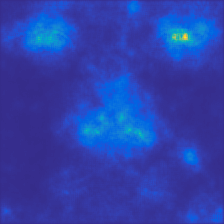}
\centering
{\small (c) Sensitivity Map}
\end{subfigure}
\begin{subfigure}[b]{0.23\textwidth}
\includegraphics[width=1\textwidth]{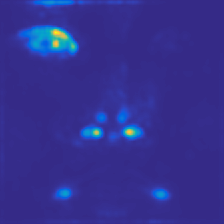}
\centering
{\small (d) FLRP}
\end{subfigure}
\begin{subfigure}[b]{0.23\textwidth}
\includegraphics[width=\textwidth]{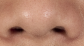}
\centering
{\small (e) magnified nose}
\end{subfigure}
\begin{subfigure}[b]{0.23\textwidth}
\includegraphics[width=1\textwidth]{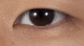}
\centering
{\small (f) magnified left eye}
\end{subfigure}
\end{center}
\caption{Example of relevance distributions (b)-(d) for a morphed face image (a) that was not detected as morph by the DNN.}
\label{fig:UseCaseExample3}
\end{figure}

If we compare the transparency masks of our evaluation framework that are generated by the three different methods, they often cover similar regions, with FLRP and LRP being more similar to each other. Figure \ref{fig:hmComparison} shows histograms of morphed face images with respect to the average cumulative differences between the transparency maps for the three interpretability methods.  These are calculated as the sum of the pixel-to-pixel differences for each pair of transparency maps generated for the evaluated methods using our framework with an $\alpha$ value of 1\%. These values are then normalized dividing them by the amount of substituted pixels.  

In a more closer inspection of the cases in which the relevance map predicted by FLRP strongly differs from those produced by LRP, we can identify another significant strength of FLRP. In such cases, the DNN is quite often wrong or uncertain in its decision. This indecision is indicated by a soft-max output for the class "morphed face image" significantly smaller than 1. Here, LRP seems to assign in most cases relevance to image regions without visible artifacts, while FLRP does still highlight regions with visible artifacts. For the morphed face image with a soft-max output smaller than 0.9 for the class morph (6.1\% of all morphed face images from the testing set), the mean normalized cumulative difference between the transparency masks for LRP and FLRP is 0.92 with a standard deviation of 0.13.

Figures \Ref{fig:UseCaseExample1},\Ref{fig:UseCaseExample2} and \Ref{fig:UseCaseExample3} show some examples of such scenarios: there is a strong dissent between LRP and FLRP and the network's decision is incorrect. This makes FLRP a suitable tool to support a human in inspecting a face image if the network does not provide a decision with a high degree of certainty, or to check where suspicious traces are in an image. All three Figures show examples of a morphed face image that is falsely classified as genuine face image. In Figure \Ref{fig:UseCaseExample1}, LRP assigns most relevance to the right eye, even though the left eye contains clearly visible morphing artifacts. The morphed face image in Figure \Ref{fig:UseCaseExample2} contains visible artifacts around the nose, but LRP assigns most of the relevance to the eyes. The relevance assignment of FLRP seems to be more reasonable in both cases. In the example shown in Figure \ref{fig:UseCaseExample3}, LRP assigns most relevance to the left eye, whereas FLRP highlights artifacts in the right eye and the nose.

\section{Summary and Discussion}
With FLRP, we introduced a method to support human inspection of possible morphed face image. Especially, in case of uncertainty in the DNN's decision, our method has shown to be able to highlight face morphing artifacts better than LRP. Additionally, we proposed a framework for the evaluation of interpretability methods for DNN-based face morphing attack detectors. 
Using this framework, we showed that the regions that are marked as relevant by the analyzed methods overlap to a certain extent. However, LRP and FLRP focus more on the cause of activation of neurons that typically respond strongly to the presence of morphs than the Sensitivity Maps do.\\
In future work, we plan to analyze the effects of image improvement~\cite{SeiboldHE19} on the detectors using our proposed framework for evaluation of interpretability methods. Moreover, we plan to extend this framework such that it can be used without having a reference image. Without having to rely on a reference image, we would also be able to use the framework for morphed face images generated by Generative Adversarial Networks~\cite{Venkatesh2020}.

\section*{Acknowledgment}
This work was partly funded by the European Union's Horizon 2020 research and innovation programme under grant agreement No 833704 (D4FLY).

{\small
\bibliographystyle{ieee_fullname}
\bibliography{FLRPArXiV_PrePub}

\begin{thebibliography}{10}\itemsep=-1pt

\bibitem{BachPLOS15}
S. Bach, A. Binder, G. Montavon, F. Klauschen, K.-R. M{\"u}ller, and W. Samek.
\newblock On pixel-wise explanations for non-linear classifier decisions by
  layer-wise relevance propagation.
\newblock {\em PLOS ONE}, 10(7):e0130140, 2015.

\bibitem{PRNUVA2}
L. Debiasi, C. Rathgeb, U. Scherhag, A. Uhl, and C. Busch.
\newblock {PRNU} variance analysis for morphed face image detection.
\newblock In {\em BTAS}, 2018.

\bibitem{London}
L. DeBruine and B. Jones.
\newblock Face research lab london set, 2017.

\bibitem{Ferrara14}
M. Ferrara, A. Franco, and D. Maltoni.
\newblock {The magic passport}.
\newblock In {\em IEEE International Joint Conference on Biometrics}, 2014.

\bibitem{Ferrara18}
M. Ferrara, A. Franco, and D. Maltoni.
\newblock Face demorphing.
\newblock {\em IEEE Transactions on Information Forensics and Security},
  13(4):1008--1017, 2018.

\bibitem{scFace}
M. Grgic, K. Delac, and S. Grgic.
\newblock Scface --- surveillance cameras face database.
\newblock {\em Multimedia Tools and Applications}, 51(3):863--879, 2011.

\bibitem{utrecht}
P. Hancock.
\newblock Utrecht ecvp, 2008.

\bibitem{PAISO}
{International Organization for Standardization}.
\newblock {ISO/IEC 30107-3:2017 Information technology -- Biometric
  presentation attack detection -- Part 3: Testing and reporting}, 2017.

\bibitem{PUT}
A. Kasiński, A. Florek, and A. Schmidt.
\newblock The put face database.
\newblock {\em Image Processing and Communications}, 13:59--64, 2008.

\bibitem{LRPRules}
M. Kohlbrenner, A. Bauer, S. Nakajima, A. Binder, W. Samek, and S. Lapuschkin.
\newblock Towards best practice in explaining neural network decisions with
  lrp.
\newblock In {\em Proceedings of the IEEE International Joint Conference on
  Neural Networks (IJCNN)}, 2020.

\bibitem{CFD}
D. Ma, J. Correll, and B. Wittenbrink.
\newblock The chicago face database: A free stimulus set of faces and norming
  data.
\newblock {\em Behavior Research Methods}, 47, 2015.

\bibitem{MakrushinND17}
A. Makrushin, T. Neubert, and J. Dittmann.
\newblock Automatic generation and detection of visually faultless facial
  morphs.
\newblock In {\em VISIGRAPP - Volume 6: VISAPP, Porto, Portugal, February 27 -
  March 1}, pages 39--50, 2017.

\bibitem{Lit2}
A. Makrushin and A. Wolf.
\newblock An overview of recent advances in assessing and mitigating the face
  morphing attack.
\newblock In {\em EUSIPCO 2018, Roma, Italy, September 3-7, 2018}, 2018.

\bibitem{Montavon2018}
G. Montavon, W. Samek, and K. M{\"u}ller.
\newblock Methods for interpreting and understanding deep neural networks.
\newblock {\em Digit. Signal Process.}, 73:1--15, 2018.

\bibitem{Neubert17}
T. Neubert.
\newblock Face morphing detection: An approach based on image degradation
  analysis.
\newblock In {\em IWDW 2017, Magdeburg, Germany, August 23-25, 2017,
  Proceedings}, pages 93--106, 2017.

\bibitem{Neubert19}
T. Neubert, C. Kraetzer, and J. Dittmann.
\newblock A face morphing detection concept with a frequency and a spatial
  domain feature space for images on {eMRTD}.
\newblock {\em In IH\&MMSec}, 2019.

\bibitem{FERET}
P.~Jonathon Phillips.
\newblock Color feret database, 2003.

\bibitem{Raghavendra16}
R. Ramachandra, K.~B. Raja, and C. Busch.
\newblock Detecting morphed face images.
\newblock In {\em BTAS}, pages 1--7, 2016.

\bibitem{Raghavendra17}
R. Ramachandra, K.~B. Raja, S. Venkatesh, and C. Busch.
\newblock Transferable deep-cnn features for detecting digital and
  print-scanned morphed face images.
\newblock In {\em CVPR Workshops}, pages 1822--1830, 2017.

\bibitem{Lit1}
U. {Scherhag}, C. {Rathgeb}, J. {Merkle}, R. {Breithaupt}, and C. {Busch}.
\newblock Face recognition systems under morphing attacks: A survey.
\newblock {\em IEEE Access}, 7, 2019.

\bibitem{Scherhag20}
U. {Scherhag}, C. {Rathgeb}, J. {Merkle}, and C. {Busch}.
\newblock Deep face representations for differential morphing attack detection.
\newblock {\em IEEE Transactions on Information Forensics and Security},
  15:3625--3639, 2020.

\bibitem{Seibold18b}
C. Seibold, A. Hilsmann, and P. Eisert.
\newblock {Reflection Analysis for Face Morphing Attack Detection}.
\newblock In {\em 26th European Signal Processing Conference (EUSIPCO)}, 2018.

\bibitem{SeiboldHE19}
C. Seibold, A. Hilsmann, and P. Eisert.
\newblock Style your face morph and improve your face morphing attack detector.
\newblock In {\em BIOSIG, Darmstadt, Germany, September 18-20}, 2019.

\bibitem{Seibold17}
C. Seibold, W. Samek, A. Hilsmann, and P. Eisert.
\newblock {Detection of Face Morphing Attacks by Deep Learning}.
\newblock In {\em IWDW 2017, Magdeburg, Germany}, 2017.

\bibitem{SeiboldSHE20}
C. Seibold, W. Samek, A. Hilsmann, and P. Eisert.
\newblock Accurate and robust neural networks for face morphing attack
  detection.
\newblock {\em J. Inf. Secur. Appl.}, 53:102526, 2020.

\bibitem{SensitivityMaps}
K. Simonyan, A. Vedaldi, , and A. Zisserman.
\newblock "deep inside convolutional networks: Visualising image classification
  models and saliency maps".
\newblock In {\em ICLR Workshop}, 2014.

\bibitem{VGG}
K. Simonyan and A. Zisserman.
\newblock Very deep convolutional networks for large-scale image recognition.
\newblock {\em CoRR}, abs/1409.1556, 2014.

\bibitem{Lit3}
S. Venkatesh, R. Ramachandra, K. Raja, and C. Busch.
\newblock Face morphing attack generation \& detection: A comprehensive survey.
\newblock abs/2011.02045, 2020.

\bibitem{Venkatesh2020}
S. Venkatesh, H. Zhang, R. Ramachandra, K. Raja, N. Damer, and C. Busch.
\newblock "can gan generated morphs threaten face recognition equally as
  landmark based morphs? - vulnerability and detection".
\newblock In {\em IWBF 2020}.

\bibitem{BU4DFE}
L. Yin, X. Chen, Y. Sun, T. Worm, and M. Reale.
\newblock A high-resolution 3d dynamic facial expression database.
\newblock In {\em 2008 8th IEEE International Conference on Automatic Face
  Gesture Recognition}, pages 1--6, 2008.

\end{thebibliography}
}

\end{document}